\documentclass[letterpaper]{article} 
\usepackage{aaai2026}  
\usepackage{times}  
\usepackage{helvet}  
\usepackage{courier}  
\usepackage[hyphens]{url}  
\usepackage{graphicx} 
\usepackage{natbib}  
\usepackage{caption} 
\usepackage{algorithm}
\usepackage{algorithmic}
\usepackage{graphicx}
\usepackage{subcaption}
\usepackage{amsmath}
\usepackage{amssymb}
\usepackage{enumitem}
\usepackage{xcolor}
\usepackage{newfloat}
\usepackage{listings}
\DeclareCaptionStyle{ruled}{labelfont=normalfont,labelsep=colon,strut=off} 
\floatstyle{ruled}
\newfloat{listing}{tb}{lst}{}
\floatname{listing}{Listing}
\title{Realistic Curriculum Reinforcement Learning for Autonomous and Sustainable Marine Vessel Navigation}
\author {
    Xiaocai Zhang\textsuperscript{\rm 1},
    Zhe Xiao\textsuperscript{\rm 2},
    Maohan Liang\textsuperscript{\rm 3},
    Tao Liu\textsuperscript{\rm 4},
    Haijiang Li\textsuperscript{\rm 5,6},
    Wenbin Zhang\textsuperscript{\rm 7}\thanks{Corresponding author.}
}
\affiliations {
    \textsuperscript{\rm 1}Department of Infrastructure Engineering, The University of Melbourne, Australia\\
    \textsuperscript{\rm 2}Institute of High Performance Computing, Agency for Science, Technology and Research (A*STAR), Singapore\\
    \textsuperscript{\rm 3}School of Navigation, Wuhan University of Technology, China\\
    \textsuperscript{\rm 4}College of Transport \& Communications, Shanghai Maritime University, China\\
    \textsuperscript{\rm 5}School of Maritime Economics and Management, Dalian Maritime University, China\\
    \textsuperscript{\rm 6}Collaborative Innovation Centre for Transport Study, Dalian Maritime University, China\\
    \textsuperscript{\rm 7}Michigan Technological University, USA\\
    xiaocai.zhang@unimelb.edu.au, xiaoz@ihpc.a-star.edu.sg, mhliang@whut.edu.cn, liutao2@shmtu.edu.cn, haijiang@dlmu.edu.cn, wenbinzh@mtu.edu
}
\usepackage{bibentry}

\begin{document}

\maketitle

\begin{abstract}
Sustainability is becoming increasingly critical in the maritime transport, encompassing both environmental and social impacts, such as Greenhouse Gas (GHG) emissions and navigational safety. Traditional vessel navigation heavily relies on human experience, often lacking autonomy and emission awareness, and is prone to human errors that may compromise safety. In this paper, we propose a Curriculum Reinforcement Learning (CRL) framework integrated with a realistic, data-driven marine simulation environment and a machine learning-based fuel consumption prediction module. The simulation environment is constructed using real-world vessel movement data and enhanced with a Diffusion Model to simulate dynamic maritime conditions. Vessel fuel consumption is estimated using historical operational data and learning-based regression. The surrounding environment is represented as image-based inputs to capture spatial complexity. We design a lightweight, policy-based CRL agent with a comprehensive reward mechanism that considers safety, emissions, timeliness, and goal completion. This framework effectively handles complex tasks progressively while ensuring stable and efficient learning in continuous action spaces. We validate the proposed approach in a sea area of the Indian Ocean, demonstrating its efficacy in enabling sustainable and safe vessel navigation.
\end{abstract}

\section{Introduction}
\label{intro}
\begin{figure}[htbp]
\begin{center}
\includegraphics[width=0.47\textwidth]{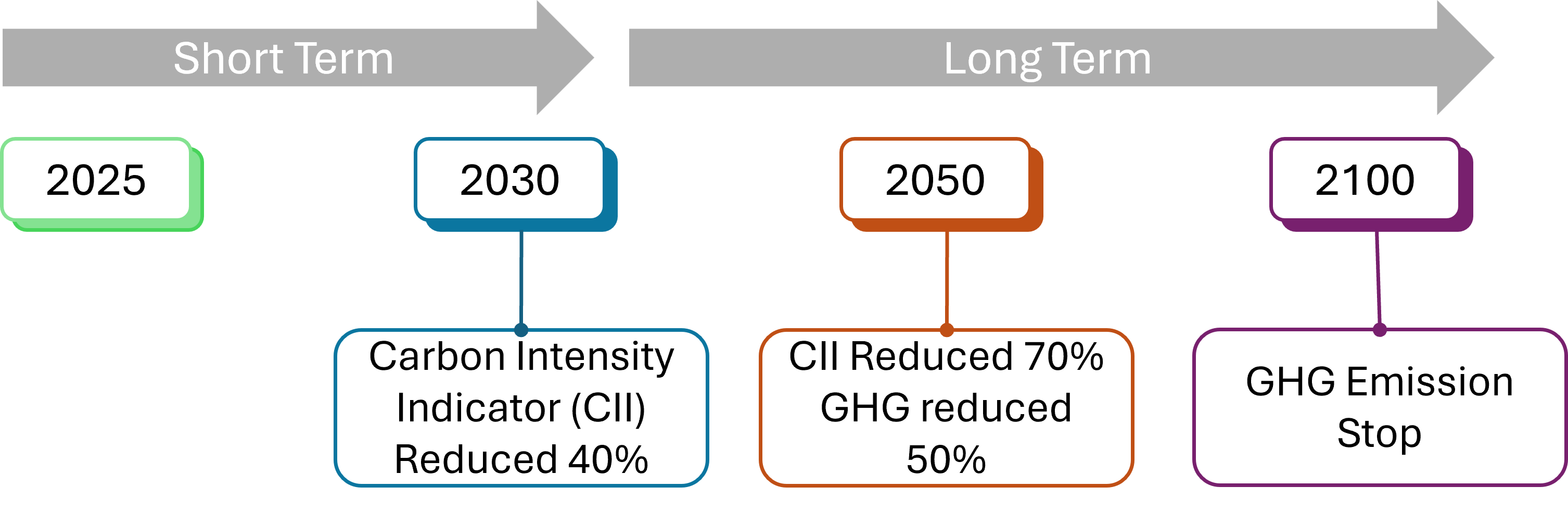}
\caption{IMO’s GHG reduction pathway.}
\label{GHG}
\end{center}
\end{figure}

The maritime transport sector plays a pivotal role in global trade and logistics, as it supports approximately 90\% of the global trade volume, being the most cost-efficient transport mode over long distances \cite{zhang2024dynamic,liu2025approach}. However, it faces growing pressure to align with sustainability goals in response to escalating environmental concerns \cite{nguyen2023blockchain,fan2024sequential,wei2025bi}. Greenhouse Gas (GHG) emissions from vessels contribute significantly to global warming. Maritime shipping produces billions of tons of GHG emissions annually. In response, the International Maritime Organization (IMO) has initiated strategies to reduce shipping emissions by at least 50\% by 2050. Despite global efforts, under a business-as-usual scenario, shipping’s carbon dioxide emissions are projected to increase by approximately 10–30\% compared to 2008 levels by 2050 \cite{huang2025review}. Figure~\ref{GHG} illustrates the IMO’s projected GHG reduction pathway.

\begin{figure*}[htbp]
\begin{center}
\includegraphics[width=0.95\textwidth]{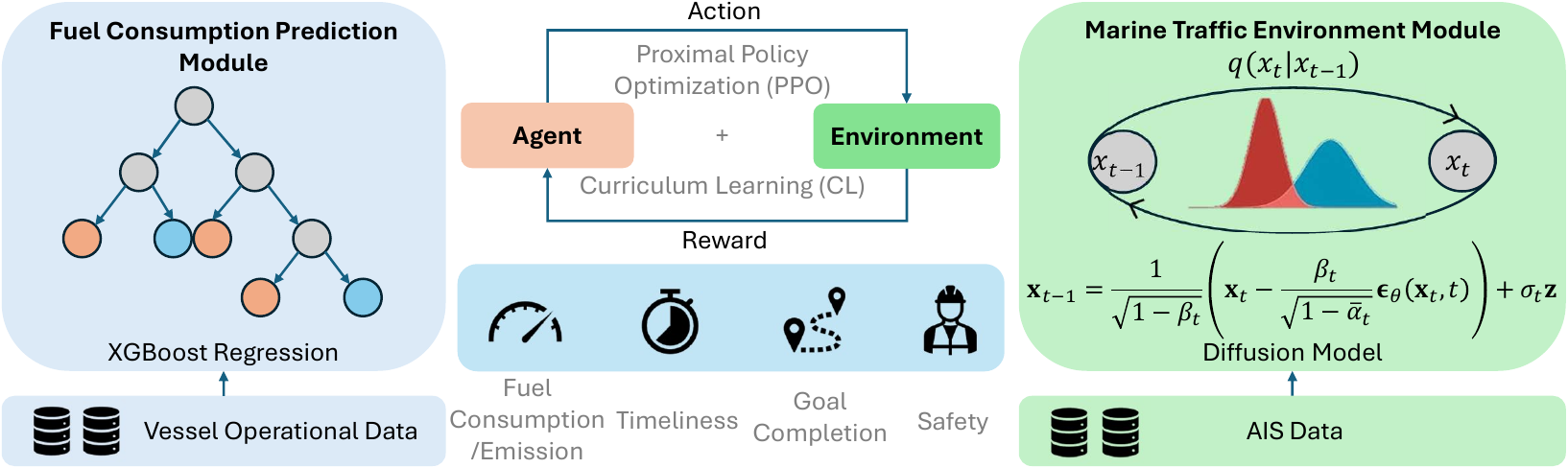}
\caption{Overview of the methodological workflow in this study.}
\label{workflow}
\end{center}
\end{figure*}

Safety represents a fundamental aspect of maritime sustainability \cite{zhou2023sustainable}. As emphasized by the IMO, a sustainable maritime transport system must not only prioritize operational efficiency and reduced emissions, but also ensure the protection of human life, cargo, and marine ecosystem by addressing risks such as oil spills, collisions, and fatalities. Failures in safety management can lead to catastrophic outcomes. One notable example is the collision between the SANCHI oil tanker and a bulk carrier in East China Sea, which resulted in the loss of 32 crew members and the discharge or combustion of more than 100,000 tons of petroleum products in the marine environment \cite{zhang2022vessel}. Investigations concluded that human error was the primary cause of the incident \cite{wan2018human}.

Optimizing marine operations provides a practical and high-impact approach to reducing GHG emissions within the maritime sector. Unlike fuel-switching or vessel retrofitting, operational measures, such as slow steaming, optimized routing, speed regulation, and Just-In-Time (JIT) arrival strategies \cite{zhang2023prediction}, can be rapidly adopted across the current global fleet with minimal capital expenditure. These strategies not only lower fuel consumption but also improve overall operational efficiency and safety, offering both environmental and economic benefits. According to the IMO, such improvements have the potential to achieve emission reductions of up to 30–40\% by 2030 \cite{comer2018relating}, serving as a crucial interim solution while low- and zero-emission technologies continue to advance. In this context, the application of advanced Artificial Intelligence (AI) techniques to maritime navigation systems emerges as a timely and promising opportunity to revolutionize conventional operational practices.

To support decarbonization in marine vessel operations, recent studies leveraging Deep Reinforcement Learning (DRL) have primarily focused on operational optimization tasks such as routing and scheduling. For example, Moradi et al. \cite{moradi2022marine} proposed a Deep Deterministic Policy Gradient (DDPG) framework to optimize vessel routing with the objective of reducing fuel consumption and CO$_{2}$ emissions. Similarly, Wang et al. \cite{wang2024fuel} applied a DDPG-based reinforcement learning approach to optimize transit routes for harbor tugboats in Singapore, taking into account local waterway characteristics, static obstacles, and dynamic environmental factors. Chen et al. \cite{chen2025intelligent} introduced a hybrid approach that integrates a traditional A$^{*}$ search algorithm \cite{xiao2024innovating} with a Double Deep Q-Network (DDQN) to enhance route planning, aiming to minimize sailing time, operational costs, and carbon emissions. Additionally, Hasanvand et al. \cite{hasanvand2020reliable} presented an advanced DQN-based strategy for onboard energy scheduling in all-electric ships, formulating a multi-objective optimization problem that balances emission reduction, cost efficiency, and system reliability.

In the context of safe navigation, most existing research focuses on real-time decision-making with the primary objective of collision avoidance. One study proposed an autonomous collision avoidance system for ships using DRL based on a Deep Q-Network (DQN), incorporating a quantitative risk assessment derived from the Closest Point of Approach (CPA) within a maneuverability-aware ship domain that adheres to COLREGs regulations \cite{wang2024deep}. Another work targeted strategy optimization for Maritime Autonomous Surface Ships (MASS) navigating through complex traffic scenarios. It introduced a risk-aware Safe Reinforcement Learning (SRL) approach built upon the Actor-Critic architecture, which jointly considers collision risk and decision reliability in the policy optimization process \cite{wang2024deep}. Furthermore, Pan et al. \cite{pan2025deep} developed a novel DRL framework utilizing a Dueling Double Deep Q-Network (DDDQN), specifically designed for multi-ship collision avoidance in dynamic and congested environments. This method advances the field by realistically modeling ship interactions and continuously generating safe maneuvering recommendations in real time.

Achieving the multi-objective goals of carbon emission reduction, navigational safety, and timely and successful task completion in autonomous vessel operations remains a significant challenge, even with the support of DRL techniques. Existing reinforcement learning approaches often face limitations in scalability and generalization, particularly under the highly dynamic conditions characteristic of real-world maritime environments. Furthermore, multi-objective optimization in DRL-based vessel autonomous navigation poses significant challenges for effective policy learning and deployment \cite{li2020deep}.
To address these limitations, we propose a novel Curriculum Reinforcement Learning (CRL) framework specifically designed for autonomous and sustainable marine vessel navigation. Our approach is grounded in a realistic, data-driven simulation environment constructed from real-world Automatic Identification System (AIS) data \cite{liang2024survey} and augmented with a Diffusion Model (DM) \cite{ho2020denoising} to simulate dynamic maritime traffic conditions. The real-time traffic conditions are encoded as image-based environmental representations, effectively capturing the spatial complexity of maritime scenarios.
In addition, we develop a machine learning-based fuel consumption prediction model trained on a large-scale dataset of real-world operational data including fuel usage records, enabling accurate estimation of fuel consumption. At the core of our DRL solution is a lightweight CRL agent guided by a multi-objective reward structure, facilitating efficient and robust policy training in complex, multi-constraint environments.
We evaluate the proposed framework in a region of the Indian Ocean, demonstrating its effectiveness in achieving safe, efficient, and environmentally sustainable navigation. By integrating curriculum learning, realistic environment simulation, and sustainability-oriented objectives, this work provides a scalable and principled pathway towards advancing AI-powered autonomy in maritime transport.

The main contributions are summarized as follows:
\begin{itemize}
    \item \textbf{A novel CRL framework for marine vessel navigation:} We propose a CRL framework that integrates curriculum learning and reinforcement learning paradigms tailored for autonomous vessel navigation, addressing safety, emissions, timeliness, and goal-oriented performance in complex ocean environments.
    
    \item \textbf{Realistic data-driven simulation environment:} We construct a high-fidelity simulation environment using real-world AIS-based vessel movement data, enhanced by a DM to simulate diverse and dynamic maritime traffic conditions.
    
    \item \textbf{Integrated fuel consumption prediction module:} We develop a data-driven fuel estimation model using historical real-world operational records from ocean-going vessels, enabling the agent to receive realistic and accurate fuel consumption feedback and make emission-aware navigation decisions.
\end{itemize}

\section{Methods}
\label{method}
\subsection{Overview}
An overview of the methodological workflow employed in this study is illustrated in Figure~\ref{workflow}. The proposed framework consists of three key components: a fuel consumption prediction module, a marine traffic environment module, and CRL-based policy learning modules. The fuel prediction module leverages real-world operational data from vessels to estimate fuel usage accurately. The marine traffic environment module simulates dynamic and realistic maritime conditions using AIS data, providing a high-fidelity training ground for policy learning. The CRL framework is driven by a multi-objective reward function that jointly optimizes fuel and emission reduction, navigational safety, timeliness, and successful task completion.

\subsection{Fuel Consumption Prediction}
Given the direct relationship between fuel usage and GHG emissions—particularly CO$_{2}$, NO$_x$, and SO$_x$—robust prediction models are critical for achieving both economic and environmental objectives. These models enable ship operators to estimate fuel requirements for specific voyages based on various dynamic factors, including vessel characteristics, engine specifications, cargo load, weather conditions, and sea state. However, due to limitations in data availability, some of these factors are challenging to obtain consistently.

\begin{figure}[htbp]
\centering
\begin{subfigure}[a]{0.47\textwidth}
    \includegraphics[width=\textwidth]{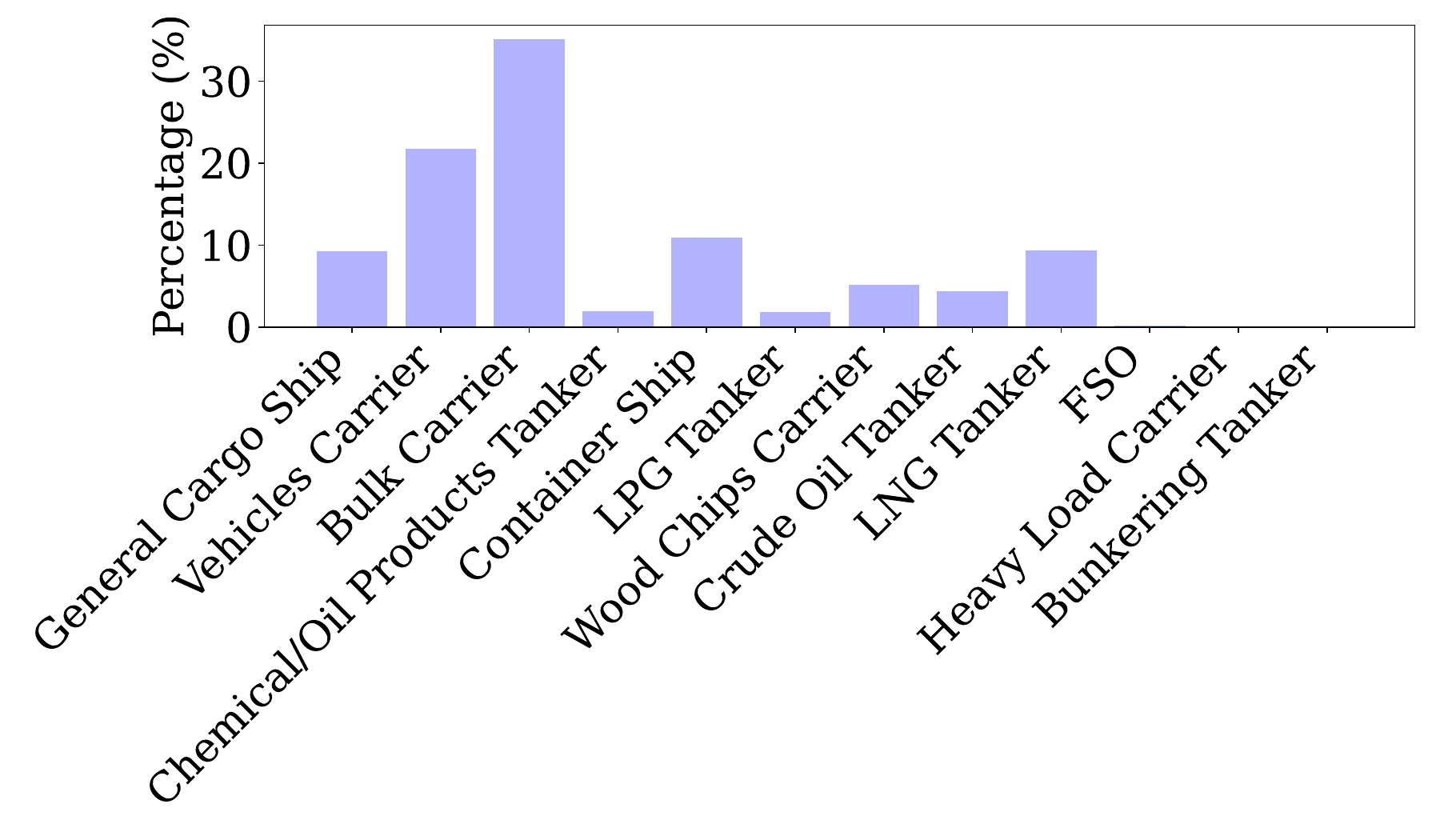}
    \caption{Ship type distribution}
    \label{fig:ship_type}
\end{subfigure}
\hfill
\begin{subfigure}[b]{0.47\textwidth}
    \includegraphics[width=\textwidth]{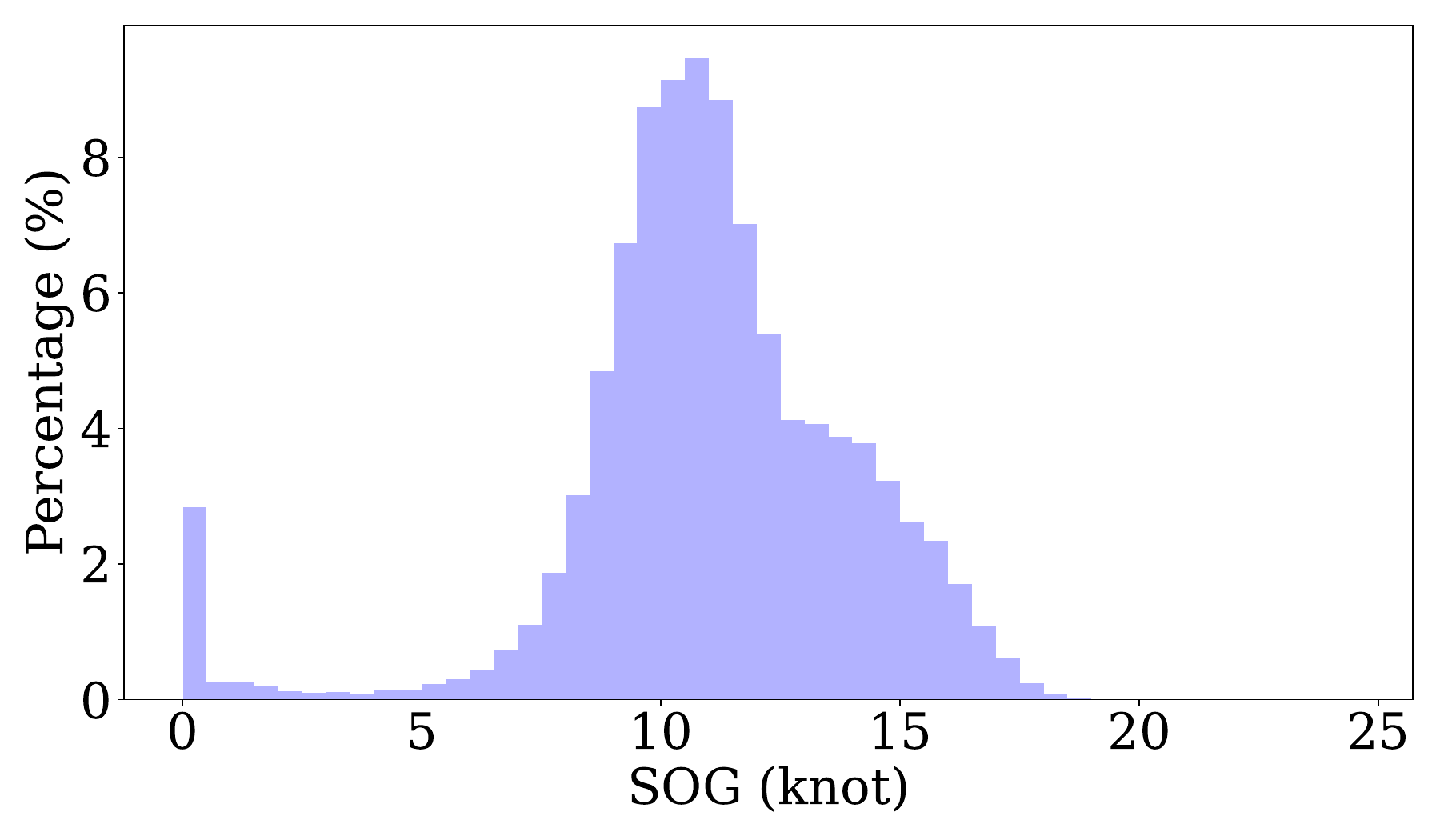}
    \caption{SOG distribution}
    \label{fig:sog}
\end{subfigure}
\caption{Distribution of vessel features}
\label{fig:distribution}
\end{figure}

To address this, we propose a machine learning-based prediction model trained on real-world operational data collected from several hundred international ocean-going vessels over a two-year period. The input features include vessel travel distance, geographical location (latitude and longitude), speed-over-ground (SOG), and vessel particulars such as Length Overall (LOA), beam, Gross Tonnage (GT), and ship type. Temporal variables, such as month, day, and hour, are also incorporated, as they are known to correlate with varying ocean and sea conditions \cite{caires2005100}. Categorical features, including ship type and temporal information, are processed using one-hot encoding. Figure~\ref{fig:distribution} shows the distributions of vessel SOG and ship types.

The model outputs the Fuel Consumption Rate (FCR), measured in metric tons (mt) of fuel per hour. In this study, fuel consumption is measured comprehensively to include usage from the main engine, auxiliary engine, and auxiliary machinery. Four types of fuel are considered: Heavy Fuel Oil (HFO), Low Sulfur Fuel Oil (LSFO), Diesel Oil (DO), and Low Sulfur Gas Oil (LSGO). The total fuel consumption represents the aggregate usage of all four fuel types.

Following insights from a recent literature review \cite{nguyen2025maritime}, we adopt eXtreme Gradient Boosting (XGBoost) to predict the FCR using the selected input features. Let the input feature vector be denoted by $\mathbf{x} \in \mathbb{R}^d$, where $d = 86$ represents the total number of features. The prediction model is formulated as:
\begin{equation}\label{eq1}
\hat{y} = f_{\text{xgboost}}(\mathbf{x}) = \sum_{k=1}^K f_k(\mathbf{x}), \quad f_k \in \mathcal{F},
\end{equation}
where $\hat{y}$ is the predicted FCR, $K$ is the number of decision trees in the ensemble, and each $f_k$ is a regression tree belonging to the functional space $\mathcal{F}$.

\subsection{DM-Enhanced Marine Traffic Environment}
To enrich the realism and variability of vessel trajectory data in the simulation environment, we incorporate a generative model \cite{zhang2024viewpoint,zhang2024dynamic} to generate synthetic AIS trajectories that mimic real-world movement patterns. Diffusion models, originally developed for image generation, have shown promising results in sequential data generation tasks due to their ability to learn complex distributions through iterative denoising.

Let a vessel trajectory be represented as a sequence of positions:
\begin{equation}\label{eq2}
\mathbf{x}_0 = ((\phi_{t,1},\lambda_{t,1},v_{t,1}), \cdots , (\phi_{t,T},\lambda_{t,T},v_{t,T})), \mathbf{x}_0\in\mathbb{R}^{T\times3},
\end{equation}
where $\mathbf{x}_0$ is the original trajectory and $T$ is the number of time steps. $\phi_{t}$, $\lambda_{t}$, and $v_{t}$ are latitude, longitude, and SOG, respectively. The forward diffusion process gradually adds Gaussian noise to the data over $N$ steps:
\begin{equation}\label{eq4}
q(\mathbf{x}_t \mid \mathbf{x}_{t-1}) = \mathcal{N}(\mathbf{x}_t; \sqrt{1 - \beta_t} \mathbf{x}_{t-1}, \beta_t \mathbf{I}), \quad t = 1, \dots, N
\end{equation}
where \(\beta_t\) is a variance schedule controlling the noise level.

During training, the model learns to reverse this process by estimating the added noise $\boldsymbol{\epsilon}$, using a neural network $\boldsymbol{\epsilon}_\theta$:
\begin{equation}\label{eq5}
\mathcal{L}_{\text{DM}} = \mathbb{E}_{\mathbf{x}_0, \boldsymbol{\epsilon}, t} \left[ \left\| \boldsymbol{\epsilon} - \boldsymbol{\epsilon}_\theta(\mathbf{x}_t, t) \right\|^2 \right]
\end{equation}

To sample a new trajectory, we initialize with Gaussian noise $\mathbf{x}_N \sim \mathcal{N}(0, \mathbf{I})$, and iteratively apply the learned reverse process, as shown by
\begin{equation}\label{eq6}
\scriptsize
\mathbf{x}_{t-1} = \frac{1}{\sqrt{1 - \beta_t}} \left( \mathbf{x}_t - \frac{\beta_t}{\sqrt{1 - \bar{\alpha}_t}} \boldsymbol{\epsilon}_\theta(\mathbf{x}_t, t) \right) + \sigma_t \mathbf{z}, \quad \mathbf{z} \sim \mathcal{N}(0, \mathbf{I}),
\end{equation}
where $\bar{\alpha}_t = \prod_{s=1}^{t} (1 - \beta_s)$ and $\sigma_t$ is the noise scale.

By leveraging the DM, we are able to generate plausible and diverse vessel trajectories that align with realistic movement patterns observed in AIS data. These synthetic trajectories enrich the simulated environment, thereby improving the quality of downstream DRL model training.

\subsection{CRL Framework}
We formulate the vessel navigation task as a Markov Decision Process (MDP), which is typically defined by four key components: state, action, state transition, and reward \cite{ni2023learning}. The curriculum learning introduces a shrinking distance-to-goal threshold that decreases with training episode.

The state representation at time step $t$ consists of two components: the self-vessel state $\mathbf{s}_1^t$ and the environmental state $\mathbf{s}_2^t$. The self-vessel state $\mathbf{s}_1^t$ captures the vessel’s status and contextual information, such as current speed and direction, at its current position. $\mathbf{s}_1^t$ is defined by
\begin{equation}\label{eq7}
\mathbf{s}_1^t = \left[\phi_t, \lambda_t, \phi_t^d, \lambda_t^d, \psi_t, \beta_t, \vartheta_t\right], \quad \mathbf{s}_1^t \in \mathbb{R}^9,
\end{equation}
where $\phi_t, \lambda_t$ denote the geographical coordinates (latitude and longitude) at time $t$, and $\phi_t^d, \lambda_t^d$ represent the destination coordinates. $\psi_t$ is the vessel's heading angle, $\beta_t$ is the direction of the ocean current, and $\vartheta_t$ is the vessel’s current speed. All angle-related features (e.g., heading and current direction) are encoded using a two-dimensional representation (e.g., sine, cosine encoding), resulting in a total state dimension of 9.

The environmental state $\mathbf{s}_2^t$ is encoded as a three-channel image tensor representing the vessel’s surrounding traffic conditions, as fromulated by
\begin{equation}\label{eq8}
\mathbf{s}_2^t \in \mathbb{R}^{64 \times 64 \times 3}.
\end{equation}

Each channel corresponds to a $64 \times 64$ grid centered on the self-vessel, where each pixel represents a discretized spatial cell in the local coordinate frame, covering a fixed area (e.g., 10 nm × 10 nm). The three channels are defined as follows:
\begin{itemize}
    \item \textbf{Channel 0}: \textit{Occupancy} — a binary indicator of whether a vessel is present in the cell.
    \item \textbf{Channel 1}: \textit{SOG} — SOG of the vessel in the cell.
    \item \textbf{Channel 2}: \textit{Course-Over-Ground (COG)} — COG of the vessel in the cell.
\end{itemize}

The action space is defined as a continuous two-dimensional space comprising the heading increment and Speed Through Water (STW). At each time step $t$, the action is represented by
\begin{equation}\label{eq9}
\mathbf{a}_t = \left[\Delta \psi_t, v_t\right], \quad -\hat{\psi} \leq \Delta \psi_t \leq \hat{\psi}, \quad v_l \leq v_t \leq v_u,
\end{equation}
\begin{equation}\label{eq10}
\psi_{t+1} = \psi_{t}+\Delta \psi_t,
\end{equation}
where $\Delta \psi_t$ denotes the change in heading angle, and $v_t$ is the commanded STW. The intervals $\left[-\hat{\psi}, \hat{\psi} \right]$ and $\left[v_l, v_u\right]$ define the allowable ranges for heading adjustment and speed, respectively. The vessel’s COG and SOG are then computed by factoring in the effects of ocean currents on the vessel’s intended motion. The subsequent position of the vessel is predicted through simulation based on the applied action and environmental conditions.

The reward function is designed to capture multiple objectives, including safety, fuel consumption, timeliness, and goal completion. Formally, the reward at time step $t$ is demonstrated by
\begin{equation}\label{eq11}
\small
r_{t} =
\begin{cases}
30 + 1.5g_{t} - f_{t} - s_{t}, & \text{if } d_{\text{cur}} < \omega(e) \\
1.5g_{t} - f_{t} - s_{t} - 1.0, & \text{if } d_{\text{cur}} > d_{\text{pre}} \\
1.5g_{t} - f_{t} - s_{t} - 0.1 \cdot d_{\text{cur}}, & \text{if late and } d_{\text{cur}} \geq \omega(e) \\
1.5g_{t} - f_{t} - s_{t}, & \text{otherwise}\\
\end{cases},
\end{equation}
where $g_{t}$, $f_{t}$, and $s_{t}$ denote the reward components for goal completion, fuel consumption, and safety, respectively. The terms $d_{\text{cur}}$ and $d_{\text{pre}}$ represent the distances from the current and previous vessel positions to the destination, respectively. In the curriculum learning paradigm, we introduce a threshold parameter $\omega(e)$ that progressively decreases over training episodes, transitioning from more challenging to simpler tasks to facilitate gradual learning. The parameter $\omega(e)$ is defined by
\begin{equation}\label{eq12}
\omega \left ( e \right )=\omega _{0}\cdot \left ( 1-\frac{e}{N_{e}} \right )+\omega_{f}\cdot \textup{min}\left ( \frac{e}{N_{e}},1 \right ),
\end{equation}
where $e$ denotes the episode index, and $\omega_0$ is the initial threshold, set to 5 nautical miles (nm) in this study.
The goal completion reward $g_t$ is calculated by
\begin{equation}\label{eq13}
g_{t} = d_{\text{pre}} - d_{\text{cur}}.
\end{equation}
The fuel consumption reward $f_{t}$ is computed by
\begin{equation}\label{eq14}
f_{t} = \alpha \cdot \frac{f_{XGBoost}(\mathbf{x}_{t})}{GT},
\end{equation}
The safety reward $s_t$ is computed based on the Distance at Closest Point of Approach (DCPA) and the Time to Closest Point of Approach (TCPA), defined as follows:
\begin{equation}\label{eq15}
\scriptsize 
s_{t} = \frac{1}{N} \sum_{i=1}^{N} 
\left[
\operatorname{clip}\left(1 - \frac{\text{DCPA}_i}{d_{\text{safe},i}},\ 0,\ 1\right)
\cdot
\operatorname{clip}\left(1 - \frac{|\text{TCPA}_i|}{t_{\max}},\ 0,\ 1\right)
\right],
\end{equation}
\begin{equation}\label{eq16}
d_{safe} = \textup{max }\left ( \tau \cdot \frac{L_{s}+B_{s}+L_{t}+B_{t}}{2\times 1852},0.5 \right ),
\end{equation}
where $L$ and $B$ means LOA and beam of a vessel, respectively. Subscripts $s$ and $t$ denote the self-vessel and target vessel involved in a potential collision scenario. To define the safety thresholds, we introduce a buffer multiplier $\tau$, set to 4 in this study. Additionally, the time threshold for collision risk assessment is capped by a maximum allowable time to closest approach, and $t_{max} = 15$ minutes.

\begin{figure}[htbp]
\begin{center}
\includegraphics[width=0.47\textwidth]{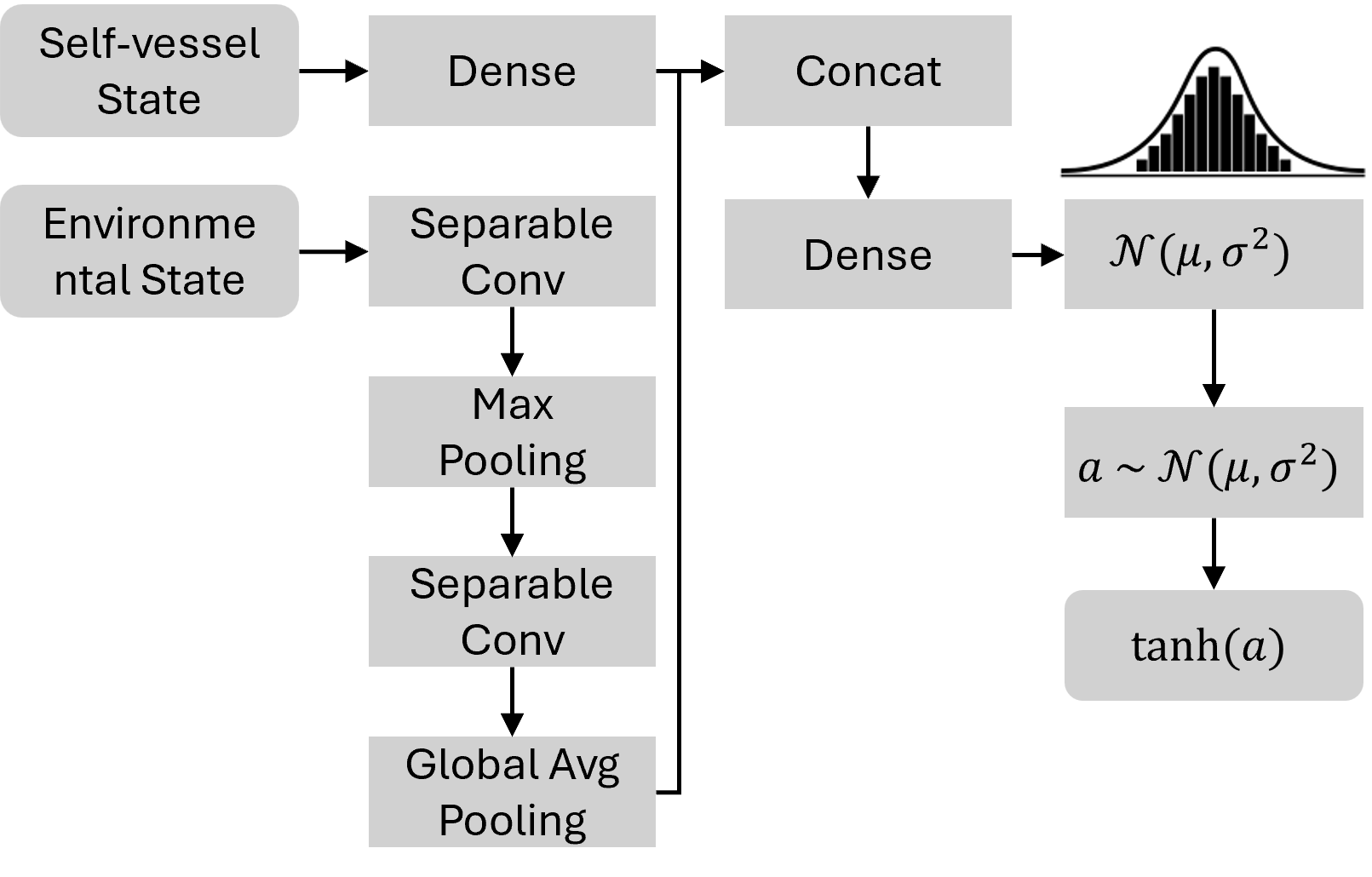}
\caption{Architecture of the actor network in CRL.}
\label{actor}
\end{center}
\end{figure}

The DRL framework in this study is built upon the Proximal Policy Optimization (PPO) algorithm, which comprises two components: an actor network and a critic network. The architecture of the actor network is depicted in Figure \ref{actor}.
The network receives two types of input states, each processed through distinct feature extraction pathways. The environmental state, represented as an image-based tensor, is processed using a lightweight convolutional backbone composed of two separable convolutional layers, allowing for efficient spatial feature extraction. In parallel, the self-vessel state, represented as a structured vector, is processed through dense layers.
The extracted features from both branches are then concatenated into a joint latent representation, which is used to compute the action distribution. The critic network adopts a similar architecture to the actor, sharing the same input processing structure, but outputs a scalar value representing the estimated value of the current state.

\section{Experiments \& Results}
\label{experiment}
The experiments are conducted through simulation within a selected sea area in Indian Ocean, chosen for its high density of commercial vessel traffic and diverse environmental conditions. While open AIS data for this region is limited, and comprehensive operational datasets for vessel fuel consumption are typically confidential and not publicly accessible, we utilize proprietary data obtained through industry collaboration. To evaluate performance under varying conditions, we design three representative navigation instances. A moderate traffic scenario is simulated, comprising hundreds of vessels distributed across the region to emulate realistic maritime operational environments.
To further enhance environmental fidelity, monthly-varying ocean currents are modeled based on historical distributions of current direction and speed. These currents introduce dynamic disturbances that influence vessel trajectories and are integrated into the simulation as time-varying environmental factors.
The combined modeling of traffic density and dynamic current flows enables a comprehensive evaluation of the proposed navigation strategy’s robustness and adaptability across a range of realistic and challenging operational conditions.

\subsection{Fuel Consumption Prediction Performance}
The prediction performance is evaluated using an independent test dataset. Three commonly used metrics—Mean Absolute Error (MAE), Root Mean Squared Error (RMSE), and the coefficient of determination (R$^2$)—are employed to assess regression model accuracy. The proposed XGBoost model is compared against several widely adopted machine learning algorithms, including Support Vector Regression (SVR), Multi-Layer Perceptron (MLP), and Random Forest (RF).
Overall, XGBoost and RF demonstrate superior performance among the evaluated models. XGBoost achieves the highest R$^2$ score of 86.10\%, indicating strong predictive capability. Furthermore, XGBoost slightly outperforms RF in terms of both RMSE and R$^2$, confirming its effectiveness for fuel consumption prediction in maritime applications.

\begin{table}[h]
\centering
\begin{tabular}{lccc}
\hline
\textbf{Method} & \textbf{MAE} & \textbf{RMSE} & \textbf{R$^{2}$} (\%) \\
\hline
SVR & 0.4529 & 0.7050 & 45.01\\
MLP & 0.2440 & 0.4916 & 77.06\\
ET & 0.2116 & 0.4033 & 84.56\\
LightGBM & 0.2015 & 0.3895 & 0.8560\\
RF & \textbf{0.1752} & 0.3832 & 86.06\\
XGBoost & 0.1802 & \textbf{0.3827} & \textbf{86.10} \\
\hline
\end{tabular}
\caption{Performance comparison between fuel consumption prediction models.}
\label{tab1}
\end{table}

\subsection{Instance Case 1}
\begin{figure}[htbp]
\begin{center}
\includegraphics[width=0.4\textwidth]{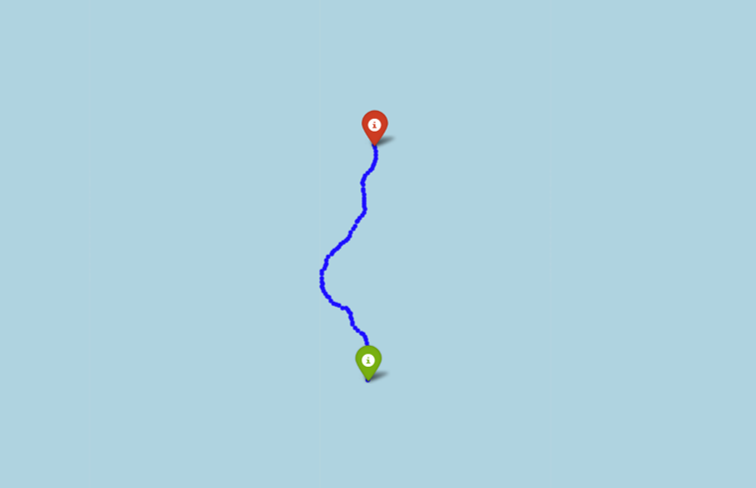}
\caption{Vessel navigation trajectory for Instance Case 1.}
\label{vis1}
\end{center}
\end{figure}

This instance simulates a vessel navigation task with a time constraint of 21 hours, spanning from 2024-02-08 00:00:00 UTC to 2024-02-08 21:00:00 UTC, requiring the vessel to travel from a specified starting point to its destination. Table \ref{tab2} presents the performance comparison of various control models based on three key metrics: Accumulated Reward (AR), Accumulated Fuel Consumption (AFC), and Accumulated Safety Score (ASS), where a higher ASS indicates greater navigational risk.
The baseline models include CL-ABDDQN, CL-A2C, and DDPG \cite{wang2024fuel,moradi2022marine}. Both CL-ABDDQN and CL-A2C adopt the curriculum learning paradigm within the Advantage Actor-Critic (A2C) framework. All baselines are trained and evaluated under the same simulated environment and reward structure as the proposed CRL framework, with the only difference being the underlying DRL architecture. The ABDDQN model builds upon the DDQN with an action-branching architecture, which effectively handles high-dimensional discrete action spaces by decoupling action components. This branching structure enhances learning stability and improves exploration efficiency \cite{tavakoli2018action}.
Among all models, the proposed CRL framework demonstrates the best overall performance across all three evaluation metrics. Notably, CL-A2C fails to complete the navigation task, resulting in a negative AR. The DDPG model underperforms, yielding lower AR and higher AFC and ASS values. While CL-ABDDQN demonstrates competitive performance, it still falls short of CRL in terms of fuel efficiency and safety. The navigation trajectory for this scenario is illustrated in Figure~\ref{vis1}, where the blue marker indicates the starting point and the red marker denotes the destination.

\begin{table}[h]
\centering
\begin{tabular}{lccc}
\hline
\textbf{Method} & \textbf{AR} & \textbf{AFC} & \textbf{ASS} \\
\hline
CL-ABDDQN & \textbf{155.527} & 20.509 & 1.466\\
CL-A2C & -82.034 & \textbf{13.092} & 7.501\\
DDPG & 150.424 & 33.533 & 1.84\\
CRL & 154.018 & 18.015 & \textbf{0.888}\\
\hline
\end{tabular}
\caption{Performance comparison of different control models in Instance Case 1.}
\label{tab2}
\end{table}

\subsection{Instance Case 2}
\begin{figure}[htbp]
\begin{center}
\includegraphics[width=0.4\textwidth]{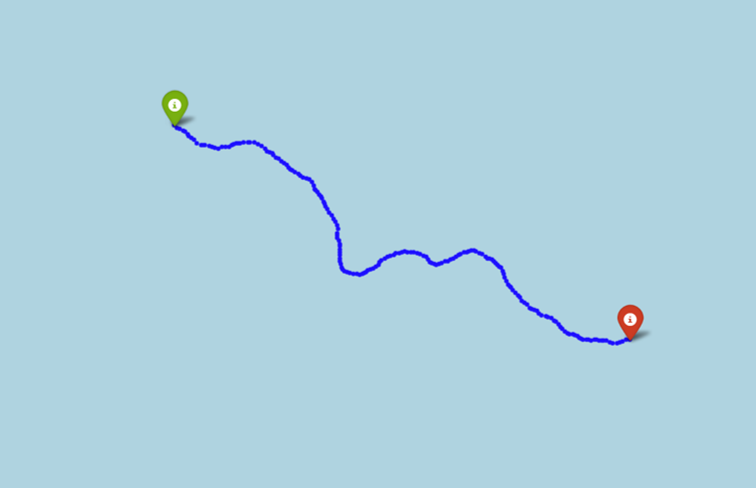}
\caption{Vessel navigation trajectory for Instance Case 2.}
\label{vis2}
\end{center}
\end{figure}

This instance simulates a vessel navigation task over a 46-hour period, from 2024-02-08 00:00:00 UTC to 2024-02-09 22:00:00 UTC, requiring the vessel to travel from a specified starting point to its destination. Table \ref{tab3} presents the performance of various control models. Due to the longer duration compared to Instance Case 1, all three metrics exhibit higher values across the models. The proposed CRL framework achieves the lowest AFC while maintaining a balanced AR and ASS. CL-ABDDQN obtains the highest reward, whereas CL-A2C yields the lowest safety score, indicating a safer trajectory. Both models incorporate the Curriculum Learning (CL) framework, highlighting its positive impact on DRL performance. Overall, CRL demonstrates the most balanced performance across all three indicators. The navigation trajectory for this instance case is visualized in Figure \ref{vis2}, where the blue marker indicates the starting point and the red marker marks the destination.

\begin{table}[h]
\centering
\begin{tabular}{lccc}
\hline
\textbf{Method} & \textbf{AR} & \textbf{AFC} & \textbf{ASS} \\
\hline
CL-ABDDQN & \textbf{298.026} & 36.229 & 5.543\\
CL-A2C & 287.175 & 21.487 & \textbf{3.952}\\
DDPG & 279.439 & 20.974	& 5.063\\
CRL & 294.148 & \textbf{19.963} & 4.754\\
\hline
\end{tabular}
\caption{Performance comparison of different control models in Instance Case 2.}
\label{tab3}
\end{table}

\subsection{Instance Case 3}
\begin{figure}[htbp]
\begin{center}
\includegraphics[width=0.4\textwidth]{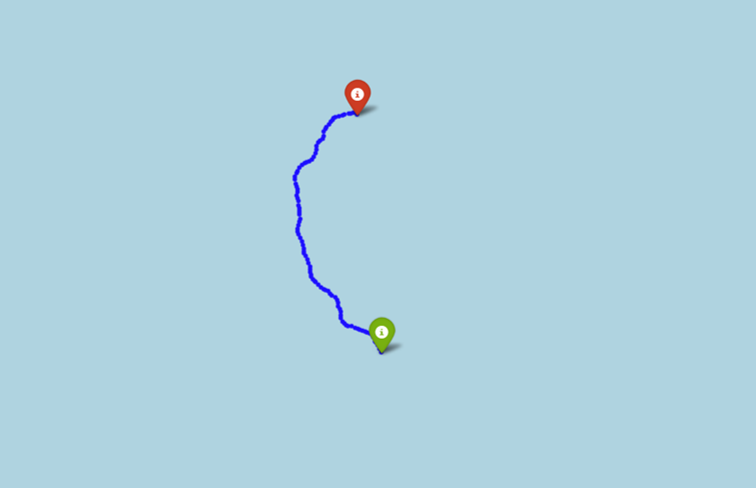}
\caption{Vessel navigation trajectory for Instance Case 3.}
\label{vis3}
\end{center}
\end{figure}

This instance simulates a vessel navigation task with a 21-hour time window, from 2024-02-08 00:00:00 UTC to 2024-02-08 21:00:00 UTC, requiring the vessel to travel from a designated starting point to its destination. Table~\ref{tab4} summarizes the performance of various control models. In this case, the CRL framework achieves the lowest ASS, indicating the safest navigation behavior, although it incurs a slightly higher AFC compared to the most fuel-efficient models. Conversely, the CL-A2C model records a negative AR (-100.699) along with high AFC and ASS values, reflecting failed goal completions and divergence in learning. The DDPG model demonstrates the best fuel efficiency, achieving the lowest AFC, but at the expense of safety, as indicated by the highest ASS.
Overall, while CRL does not significantly outperform all baseline models, it shows a clear advantage over CL-A2C. Its performance is comparable to that of CL-ABDDQN, with both models achieving a balanced trade-off among reward, safety, and fuel consumption. Figure~\ref{vis3} illustrates the vessel’s trajectory for this scenario, where the blue marker indicates the starting point and the red marker represents the destination.

\begin{table}[h]
\centering
\begin{tabular}{lccc}
\hline
\textbf{Method} & \textbf{AR} & \textbf{AFC} & \textbf{ASS} \\
\hline
CL-ABDDQN & \textbf{155.035} & 20.901 & 2.332\\
CL-A2C & -100.699 & 35.154 & 7.258\\
DDPG & 148.615 & \textbf{9.002} & 2.710\\
CRL & 145.015 & 28.618 & \textbf{2.143}\\
\hline
\end{tabular}
\caption{Performance comparison of different control models in Instance Case 3.}
\label{tab4}
\end{table}

\subsection{Ablation Analysis}
To evaluate the impact of the proposed curriculum learning paradigm within the DRL framework, we perform an ablation analysis by comparing the full model (CRL) with a variant that excludes the CL component (CRL w/o CL). The CRL model incorporates curriculum learning into the training process, while CRL w/o CL shares the same architecture but is trained without the CL mechanism. The results are summarized in Table~\ref{ablation}.
In Instance Case 2, the CRL w/o CL model obtains a negative AR, indicating failure to complete the navigation task and highlighting the critical role of curriculum learning in guiding the agent toward successful goal completion under challenging conditions. In Instance Cases 1 and 3, the full CRL model outperforms its counterpart across all evaluation metrics, with the exception of fuel consumption in Instance Case 3, where CRL w/o CL shows a slight advantage in fuel efficiency. These results confirm the effectiveness and necessity of curriculum learning in enhancing training stability, task success rate, and overall performance in complex maritime navigation scenarios.

\begin{table}[h]
\centering
\begin{tabular}{llccc}
\hline
\textbf{Instance} & \textbf{Method} & \textbf{AR} & \textbf{AFC} & \textbf{ASS} \\
\hline
Case 1 & w/o CL & 139.629 & 21.893 & 1.069 \\
       & CRL & \textbf{154.018} & \textbf{18.015} & \textbf{0.888} \\
Case 2 & w/o CL & -708.765 & 48.951 & \textbf{3.274} \\
       & CRL & \textbf{294.148} & \textbf{19.963} & 4.754 \\
Case 3 & w/o CL & 149.950 & \textbf{19.557} & 5.901 \\
       & CRL & \textbf{145.015} & 28.618 & \textbf{2.143} \\
\hline
\end{tabular}
\caption{Ablation analysis comparing CRL with and without CL across three instance cases.}
\label{ablation}
\end{table}

\begin{figure}[htbp]
\begin{center}
\includegraphics[width=0.47\textwidth]{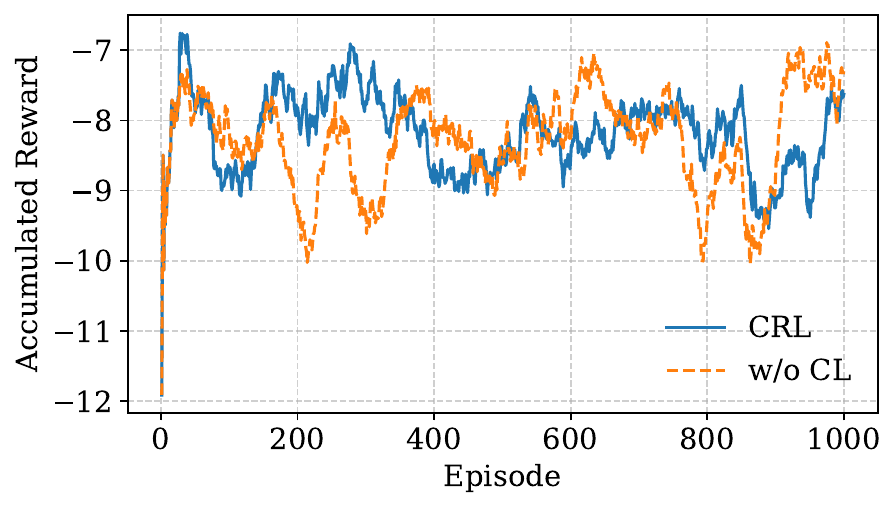}
\caption{Comparison of training reward trends between CRL and CRL without CL.}
\label{reward}
\end{center}
\end{figure}

Figure~\ref{reward} illustrates the training reward trends of the CRL model compared to its variant without the curriculum learning paradigm. The comparison reveals that CRL leads to faster convergence and greater training stability. With CRL, the agent rapidly achieves higher accumulated rewards in the early episodes and maintains stable performance throughout training, indicating efficient policy learning. In contrast, the model without curriculum learning shows slower initial improvement and more pronounced fluctuations, including periods of significant performance degradation, suggesting difficulty in exploring sparse or complex reward spaces. Overall, incorporating curriculum learning improves sample efficiency, stabilizes the learning process, and results in more consistent reward optimization.

\section{Conclusion}
\label{conclusion}
In this study, we proposed a realistic Curriculum Reinforcement Learning (CRL) framework designed for autonomous and sustainable navigation of marine vessels. The framework incorporates a data-driven simulation environment built from real-world AIS data, a generative diffusion model to capture dynamic maritime conditions, and a machine learning-based fuel consumption prediction module to provide realistic emission feedback. The DRL agent is embedded within the curriculum learning paradigm and is guided by a multi-objective reward function that jointly optimizes navigational safety, fuel efficiency, timeliness, and successful task completion, enabling efficient and stable policy learning in complex continuous control environments. Extensive simulation experiments conducted in a selected sea region of the Indian Ocean validate the effectiveness of the proposed approach. Our CRL framework demonstrates superior performance over state-of-the-art DRL baselines, consistently achieving better trade-offs among cumulative reward, safety, and fuel consumption. Additional ablation studies highlight the critical role of curriculum learning in enhancing training stability and accelerating convergence.

\section{Acknowledgments} 
Haijiang Li was supported in part by the National Natural Science Foundation of China under grant no. 52501430, and Tao Liu was supported in part by the National Natural Science Foundation of China under grant no. 52571408.

\bibliography{aaai2026}

\end{document}